\documentclass[10pt,twocolumn,letterpaper]{article}

\usepackage{wacv}     

\usepackage{graphicx}
\usepackage{amsmath}
\usepackage{amssymb}
\usepackage{booktabs}
\usepackage{adjustbox}
\usepackage{multirow}
\usepackage{todonotes}
\usepackage{float}

\usepackage[pagebackref,breaklinks,colorlinks]{hyperref}

\usepackage[capitalize]{cleveref}
\crefname{section}{Sec.}{Secs.}
\Crefname{section}{Section}{Sections}
\Crefname{table}{Table}{Tables}
\crefname{table}{Tab.}{Tabs.}

\def\wacvPaperID{6} % *** Enter the WACV Paper ID here
\def\confName{WACV}
\def\confYear{2025}

\begin{document}

\title{Enhancing Ground-to-Aerial Image Matching for Visual Misinformation Detection Using Semantic Segmentation}

\author{Emanuele Mule\thanks{First three authors contributed equally.} ,
Matteo Pannacci$^*$,
Ali Ghasemi Goudarzi$^*$,\\
Francesco Pro,
Lorenzo Papa,
Luca Maiano,
Irene Amerini\\
Sapienza University of Rome\\
{\tt\small \{mule.1939852, pannacci.1948942, ghasemigoudarzi.2053156\}@studenti.uniroma1.it}\\
{\tt\small \{pro, papa, maiano, amerini\}@diag.uniroma1.it}
% \and
% Matteo Pannacci\\
% Sapienza University of Rome\\
% % First line of institution2 address\\
% {\tt\small pannacci.1948942@studenti.uniroma1.it}
% \and
% Ali Ghasemi\\
% Sapienza University of Rome\\
% {\tt\small ghasemigoudarzi.2053156@studenti.uniroma1.it}
}
\maketitle

%%%%%%%%% ABSTRACT
\begin{abstract}
%%%% OLD 
%The recent development of generative AI techniques, which has increased the online diffusion of altered images and videos, raises concerns about the credibility of digital media accessible on the Internet and shared by information channels and social networks. Domains that rely on this data, such as journalism, forensic analysis, and earth observation, suffer because of this problem. At this aim, being capable of geolocating a non-geo-tagged ground-view image without external information, like GPS coordinates, is becoming crucial. 
%This study addresses the challenge of linking a ground-view image, which can have different field of view (FoV) values, to its corresponding satellite image without relying on GPS data. A novel four-stream Siamese-like architecture, Quadruple Semantic Align Net (SAN-QUAD), is proposed to achieve this. SAN-QUAD expands previous SOTA methods, leveraging semantic segmentation applied to both ground and satellite images. The obtained results on a subset of the CVUSA dataset show notable improvements, up to $9.8 \%$, over previous methods when tested across different FoV values.
%%% NEW
%The recent advancements in generative AI techniques, 
The recent advancements in generative AI techniques, which have significantly increased the online dissemination of altered images and videos, have raised serious concerns about the credibility of digital media available on the Internet and distributed through information channels and social networks. This issue particularly affects domains that rely heavily on trustworthy data, such as journalism, forensic analysis, and Earth observation. To address these concerns, the ability to geolocate a non-geo-tagged ground-view image without external information, such as GPS coordinates, has become increasingly critical. This study tackles the challenge of linking a ground-view image, potentially exhibiting varying fields of view (FoV), to its corresponding satellite image without the aid of GPS data. To achieve this, we propose a novel four-stream Siamese-like architecture, the Quadruple Semantic Align Net (SAN-QUAD), which extends previous state-of-the-art (SOTA) approaches by leveraging semantic segmentation applied to both ground and satellite imagery. Experimental results on a subset of the CVUSA dataset demonstrate significant improvements of up to 9.8\% over prior methods across various FoV settings.
\end{abstract}

%At the forefront of Artificial Intelligence innovation, Generative AI is changing the creation of digital content and reshaping the boundaries of technology.
%%%%%%%%% BODY TEXT
\section{Introduction} \label{sec:intro}
At the forefront of Artificial Intelligence innovation, Generative AI is changing the creation of digital content and reshaping the boundaries of technology. It represents a range of techniques that, by learning from vast datasets of images, text, sounds, or other forms of data, generate new, realistic, and high-quality content. These techniques enable even non-expert users to generate realistic fake images and videos\cite{10156981}.
The rapid proliferation of generative AI applications across the Internet has significantly amplified the spread of fake news, raising serious concerns about the credibility of online media accessible daily by everyone through a smartphone\cite{huschens2023trustchatgptperceived, loth2024blessingcursesurveyimpact, Raman2024}.

Many times altered images have been used to influence public perception and decision-making processes, as happened when Russia painted fake fighter jets at its airfield in 2024 \cite{businessinsiderRussiaPainted} or when a fake image of Europe from space was passed off as an image taken by NASA\cite{businessinsiderTheseUnbelievable}.

\begin{figure}[t]
    \centering
    \includegraphics[width=1.0\linewidth]{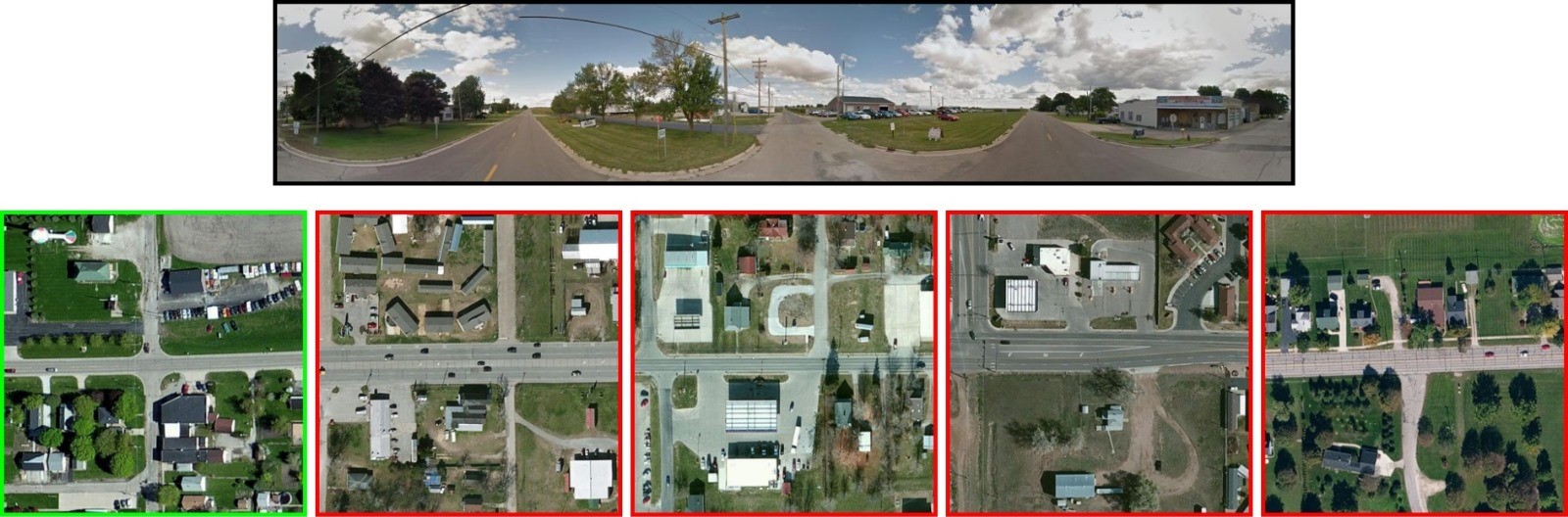}
    \caption{Example of Ground-to-Aerial Matching: This illustration demonstrates the task of identifying the corresponding satellite image for a given ground-view query. The top rectangle displays the ground-view query image while the bottom section shows the top-5 (from left to right) satellite image matches. The correct match is highlighted in green, while incorrect matches are marked in red.}
    \label{fig:enter-label}
\end{figure}

This phenomenon poses significant challenges for sectors such as journalism, forensic analysis, and Earth observation, which increasingly depend on satellite and ground-view imagery to support their activities. Consequently, verifying the origin and authenticity of non-geo-tagged images has become essential for fighting misinformation and safeguarding the integrity of visual data. One potential solution to handle this issue could be to geolocate ground-level images linking them to their corresponding satellite images without relying on external metadata such as GPS coordinates. However, this task is highly challenging, primarily due to the significant perspective discrepancies between ground-view and satellite images. Additionally, aspects such as image quality, weather conditions, seasonality, and landscape changes further aggravate the complexity of the problem.

In this context, we propose a novel approach to image geolocation aimed at improving the ability to identify manipulated visual content used for spreading misinformation. To achieve this, we introduce the Quadruple Semantic Align Network (SAN-QUAD), a four-stream Siamese-like model capable of extracting and correlating both visual and semantic features from ground-view and satellite images. Unlike state-of-the-art methods like\cite{pro2024, shi2020ilookingatjoint}, SAN-QUAD integrates semantic segmentation maps for both ground-level and satellite images, improving the ability to identify consistent features across differing perspectives.
Figure \ref{fig:enter-label} gives a graphical idea of the matching between an aerial image and a query ground-view image.

The proposed methodology demonstrates significant improvements over prior state-of-the-art approaches in image geolocation, contributing to the advancement of tools for detecting and mitigating misinformation in critical scenarios such as journalism, Earth Observation, and forensic investigation. Therefore, the contributions of this work are the following: (I) extraction of novel features from ground-view and satellite images to enhance semantic understanding, (II) introduction of a new architecture, SAN-QUAD, that combines semantic segmentation masks with RGB images, (III) use of a new strategy for merging features extracted from the different branches of the model. 

\section{Related Works}

In the early stages, the main line of research on cross-view image matching focused on generating hand-crafted features\cite{manual1,manual2}. However, these methods struggled to bridge the significant differences in point-of-view between ground-level and aerial images. 
Building on the success of deep learning in computer vision tasks, Workman and Jacobs \cite{Jacob} were the first to apply CNN-extracted features to cross-view image matching and retrieval, demonstrating its superiority over traditional hand-crafted features. On the other hand, Verde et al.\cite{9170787} and subsequently, Bonaventura et al.\cite{10167940} created a non-end-to-end deep learning approach based on generating and matching feature graph representations both for ground and aerial-view images. Despite previous research, many works using advanced Deep Learning techniques have emerged, significantly improving performance in this area.

These techniques can be grouped into three main categories: (I) \textit{Siamese-like Networks}, (II) \textit{Generative techniques}, (III) \textit{Vision Transformers}. Using a CNN architecture, Siamese-like networks aim to extract and learn viewpoint-invariant discriminative features from satellite and ground-view images. Moreover, in this category, a subset of works focused on spatially-aware image features\cite{shi2019optimalfeaturetransportcrossview, shi2020ilookingatjoint,pro2024}. More advanced methods have employed generative models, i.e. Generative Adversarial Networks (GANs)\cite{goodfellow2014generativeadversarialnetworks}, to synthesize aerial images from ground-view data and reduce the domain gap between ground and aerial images, introducing also additional information, enhancing the retrieval tasks\cite{regmi2018,deng2019,toker2021}. With the advent of Transformers\cite{vaswani2023attentionneed} and subsequently of Vision Transformers\cite{dosovitskiy2021imageworth16x16words}, the latter has been applied for this task, dividing images into patches and using the attention mechanism to capture relationships. This approach helps the model understand spatial patterns and long-range dependencies, making it well-suited for the task \cite{zhu2022, yang2021, zhang2022}.

%
%The proposed method is built on top of the SAN architecture proposed by \cite{pro2024}. The latter is based on a three-branches architecture not including as input the semantic segmentation mask for the ground-view images. The combination of the extracted aerial feature maps is done by concatenating them on the channel axis.
%
Building upon the Siamese-like Networks family, the proposed approach, which extends the methodology in Pro et al.\cite{pro2024}, integrates additional semantic segmentation masks for ground and satellite images. It introduces an additional feature extractor branch and a new feature combination methodology. Feature maps are first extracted from the ground and satellite-view images and their semantic segmentation masks, then combined to obtain global satellite
features, merging the features from the aerial images and their segmentation masks, and global ground features, combining the features from the ground-view images and their segmentation masks. The two resulting feature maps are then correlated computing a similarity matching score.

Recently, Wang et al.\cite{10.1145/3664647.3681431} exploit the combination of CNNs and Transformers to improve the performances, but opposite to the presented work, without taking into account the generalization on different FoV values. Tackling the problem from a different point of view Fan et al.\cite{s24123719} use images captured from drones instead of ground-view images, like the ones used in the presented work, aiming at geolocate the area depicted in the image taken from a drone through a matching with a satellite image depicting the same location.

\section{Proposed Method}\label{Proposed Method}

%SAN-QUAD architecture leverages semantic segmentation masks for both ground and satellite images. These segmentation features, extracted through the use of a state-of-the-art technique, provide the proposed model with contextual information to plain RGB images facilitating the matching between the two different views. Additionally, we introduce a new strategy for combining the feature maps produced by each branch of the architecture. 
%In the following sections we describe the proposed pipeline to tackle the ground-to-aerial-matching task starting from the generation of the segmentation features up to the detailed description of the designed architecture.
The SAN-QUAD architecture incorporates semantic segmentation masks for both ground and satellite images. These segmentation features, whose extraction is detailed in Sections \ref{sec311} and \ref{sec312}, enhance the contextual understanding of the model about plain RGB images, facilitating the matching process between the two distinct viewpoints. Furthermore, we introduce a novel strategy for integrating the feature maps generated by each branch of the architecture, improving its overall performance.

The subsequent sections provide a comprehensive description of the proposed pipeline, represented in Figure \ref{fig:model architecture}, for addressing the ground-to-aerial matching task. This includes details on the generation of segmentation features and an in-depth explanation of the architecture's design.

\subsection{Data Generation} \label{data generation}
The dataset used is a subset of the CVUSA dataset\cite{workman2015wideareaimagegeolocalizationaerial,zhai2016predictinggroundlevelscenelayout, pro2024}. To incorporate additional segmentation features, the dataset is enriched\footnote{The enriched dataset is available \href{https://drive.google.com/file/d/11DR7zhd6wchdyt8DSkTY2JGgf_jrtf1D}{here}}, for the aim of this work, by adding segmentation features obtained from the original ground RGB images, through Mask2Former model\cite{cheng2022maskedattentionmasktransformeruniversal}. 
Each sample of the new dataset consists of 4 different types of data: (I) Ground RGB Image, (II) Ground Semantic Segmentation Mask (III) Satellite RGB Image (IV) Satellite Semantic Segmentation Mask.

\subsubsection{Satellite Semantic Segmentation}
\label{sec311}
\begin{figure}[ht]
    \centering
    \includegraphics[width=1\linewidth]{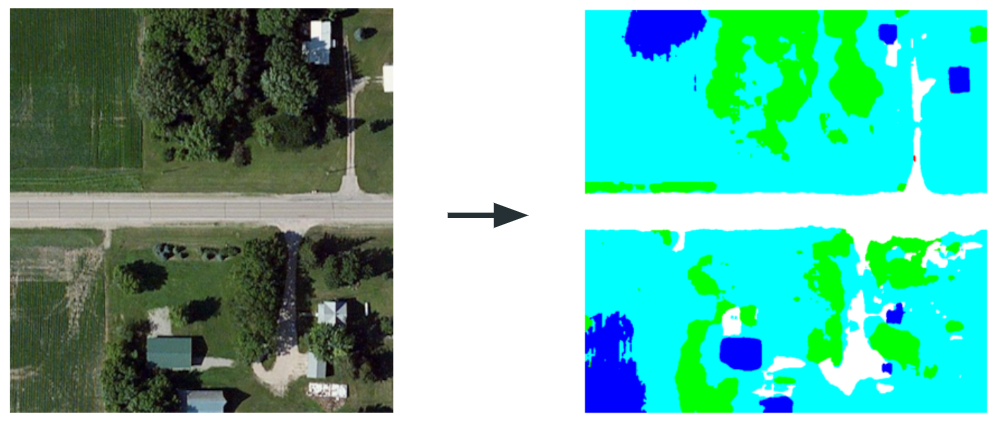}
    \caption{Example of a satellite semantic segmentation mask produced by the NEOS model.}
    \label{fig:sat_segmap}
\end{figure}

The semantic segmentation masks of satellite images have been created and employed in previous work\cite{pro2024}. Those have been generated through the use of NEOS\cite{dionelis2024}, a Transformer-based model built upon SegFormer \cite{segformer}. This model distinguishes 6 different classes: (I) high vegetation (represented with the colour green), (II) buildings (blue), (III) low vegetation (cyan), (IV) roads (white), (V) cars (yellow), and (VI) clutter (red). An example of these images is depicted in Figure \ref{fig:sat_segmap}. The same model and standard colors have been used for this work.

\subsubsection{Ground Semantic Segmentation}
\label{sec312}
\begin{figure}[ht]
    \centering
    \includegraphics[width=1\linewidth]{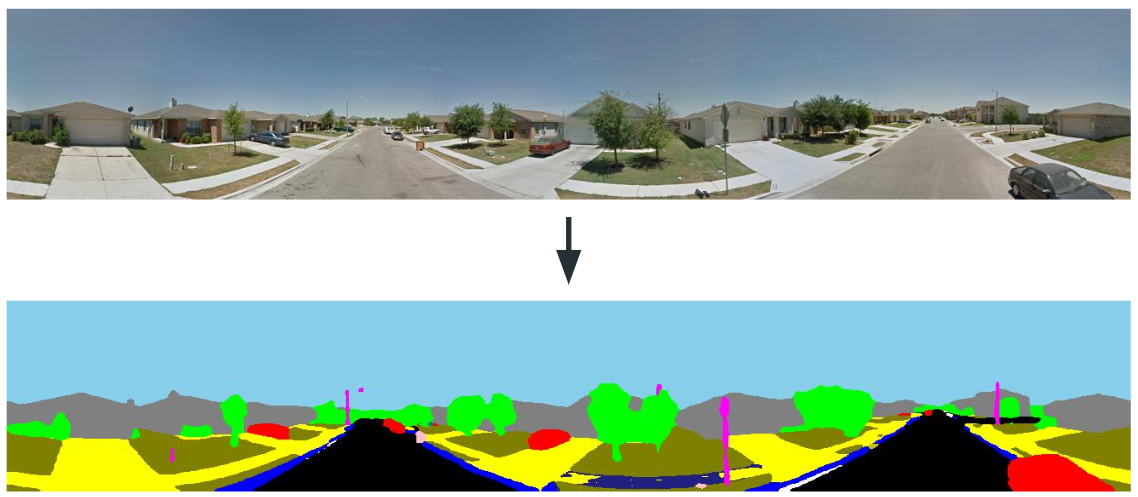}
    \caption{Example of a ground semantic segmentation mask produced by the Mask2Former model.}
    \label{fig:grd_segmap}
\end{figure}

To generate ground semantic segmentation images, the Mask2Former model\cite{cheng2022maskedattentionmasktransformeruniversal} was employed. This architecture first extracts feature maps from input images using a backbone and then it refines them through a pixel decoder to produce high-resolution features. Lastly, these features are fed into a Transformer decoder, producing a set of binary masks and class predictions. This model is pre-trained on the Mapillary Vistas Dataset\cite{MVD}. The resulting semantic segmentation masks retain the same dimensions as the original RGB images, $1232 \times 224$. While the model originally produced many classes, some were unnecessary for the treated task. To address this, a custom color palette was developed to simplify the masks by merging less significant distinctions. For instance, categories such as road, emergency lane, and bike lane, separately identified by the model, were unified into a single category named Road and represented by a single colour. This simplification allowed the segmentation process to focus on the most relevant features. An example is illustrated in Figure \ref{fig:grd_segmap}.
%\footnote{A detailed description of the ground semantic segmentation color palette can be found in the model's implementation \href{https://github.com/MatteoPannacci/SemanticAlignNet-QUAD/blob/main/feature_enrichment/ground_semantic_seg.py}{here}.} 

%\subsubsection{Ground Depth Estimation}

%Expanding on the idea of incorporating additional information into the model, the depth estimation mask of the ground viewpoint was introduced as a new type of feature mask. By explicitly providing depth information, it was anticipated that the model could better understand spatial relationships, but the results showed worse performances (see \ref{ablation}). We include these results as they may offer valuable insights for future research focused on utilizing depth estimation, specifically in the employed strategy for combining the features. To generate these depth estimation masks, the Depth-Anything model \cite{yang2024depthanythingunleashingpower} was used, specifically the \href{https://huggingface.co/LiheYoung/depth_anything_vitl14}{vitl\_14} version, which is based on Vision Transformers. As illustrated in Figure \ref{fig:grd_depmap}, the depth estimation mask is represented as an image with the same dimensions as the original one ($1232 \times 224$). A color map is applied to the normalized logits produced by the model, with the color scaling from yellow for the nearest points to black for the farthest ones.

%\begin{figure}[ht]
%    \centering
%    \includegraphics[width=1\linewidth]{imgs/grd_depmap.png}
%    \caption{Example of a ground depth estimation mask produced %by the Depth-Anything model.}
%    \label{fig:grd_depmap}
%\end{figure}

\subsection{Architecture} \label{architecture}
\begin{figure*}[ht]
    \centering
    \includegraphics[width=0.9\textwidth]{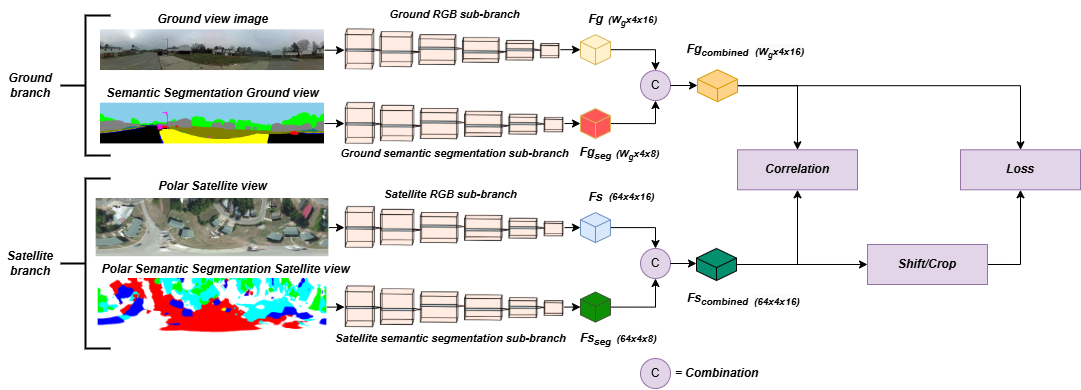}    \caption{\textbf{General overview of our methodology: SAN-QUAD.} The architecture is composed of four branches, two for the ground viewpoint and two for the satellite one. Each branch takes as input either an RGB image or a semantic segmentation mask and produces a feature volume. The volumes relative to the same viewpoint are then combined into the final feature representations which are compared to obtain the most likely orientation and perform the matching.}
    \label{fig:model architecture}
    
\end{figure*}
Figure \ref{fig:model architecture} gives an overview of the presented architecture. The SAN-QUAD model is based on four VGG16 stream branches receiving four different images as input: the ground and satellite-view images and their corresponding semantic segmentation masks. A polar transformation is applied to the two aerial-view images: the original RGB one, and its segmentation mask, as in\cite{pro2024}. This transformation mitigates the disparities between the aerial and ground perspectives, effectively aligning them within a comparable domain. It allows the assignment of an azimuth angle to each column of the aerial images, making them easily comparable with the ground views on the horizontal axis. 

Each input image has a specific network branch designed to extract features from them. The resulting feature maps are grouped and combined. Aerial features and ground features are processed separately, as illustrated in Figure \ref{fig:model architecture}. The proposed method introduces a novel strategy for combining feature maps: a partial summation between a sub-portion of them is performed. This approach is elaborated further in Section \ref{comb_strategies}. The resulting combined features are then correlated along the azimuth angle. It corresponds to assigning a similarity matching score to each possible orientation alignment between the aerial and ground-view, in this way the orientation angle between them is estimated and a maximum score corresponds to a matching between the images. The orientation estimation and the assignment of a similarity score refer to the correlation block depicted in Figure \ref{fig:model architecture} and it will be detailed in Section \ref{sim_matching}.

The proposed SAN-QUAD enables a more coherent comparison between ground and aerial features extracted from the two perspectives, overcoming the limitations of other approaches, and resulting in a significant performance improvement, achieving up to a 9.8\% increase in top-1 accuracy, as will be shown in Section \ref{comparison}

\subsubsection{Combination Strategies}
\label{comb_strategies}
The SAN-QUAD model can be conceptually divided into two horizontal components. (I) The ground: two branches extract features from the ground-view images and their associated semantic segmentation masks. (II) The satellite: the remaining two branches process features from the satellite-view images and their corresponding semantic segmentation masks.

The method used to combine feature maps into a unified representation, is a critical consideration in architectures having multiple branches. Comparing two different feature maps, both obtained with a concatenation along the channel axis may introduce limitations: having two branches for the ground-view images and two for the satellite-view ones, the result of the concatenation is that the features extracted from the RGB branch of the ground will be compared only with those extracted from the RGB branch of the satellite and the same for the segmentation, thus making the RGB and segmentation part not interact to each other.

To address this issue, we propose a shared features strategy among the branches. This is achieved by summing the feature maps derived from the semantic segmentation masks with an equivalent number of channels from the RGB feature maps while keeping the remaining RGB channels unchanged. For example, referring to Figure \ref{fig:model architecture}, \(F_{g}\) feature map has 16 channels and \(F_{g_{seg}}\) has 8 channels. The proposed combination approach produces a final representation, \(F_{g_{combined}}\), with 16 channels, in which the first 8 channels are the summation between the first 8 channels of \(F_{g}\) and all the channels of \(F_{g_{seg}}\), and its 8 left channels are equal to the last 8 channels of \(F_{g}\). Setting the number of channels for the RGB images greater than the corresponding for the segmentation mask images proved to be more effective in terms of performance.

%A way to mitigate this problem is to create shared features between the branches. We achieved this by summing the feature maps derived from the semantic segmentation masks with a corresponding number of channels from the RGB feature representation while leaving the remaining RGB channels unchanged\footnote{Here we are assuming that the number of RGB channels is greater than the number of segmentation mask channels.}. For instance, consider an example with 16 RGB feature channels and 8 semantic segmentation feature channels. The proposed combination strategy would yield a final representation comprising 16 channels: the first 8 channels contain shared features, resulting from the summation, while the last 8 channels retain RGB-only features.

This approach enhances the discriminative capacity of the resulting features, as they apply to both branches, effectively reducing the discrepancy between the two viewpoints. By integrating shared and RGB-specific features, the model achieves a more robust and coherent comparison across the two perspectives.

\subsubsection{Similarity Matching and Orientation Estimation}
\label{sim_matching}
To correctly match the spatially-aware feature representations of the two viewpoints it is necessary to align them. This task is particularly critical when dealing with limited FoV for the ground images. The polar transformation applied to the satellite images, allows the assignment of an azimuth angle to each image, making them easily comparable with the ground view over the horizontal axis. The correlation between the ground and aerial features is then computed along the azimuth angle axis using the ground features as a sliding window on top of the satellite ones. The most likely orientation will be given by the angle maximizing the correlation for that image pair. The match of a ground query image is finally obtained by taking the satellite image with the highest similarity score in the dataset.
The similarity score is calculated as:
\begin{equation}
    [F_a \ast F_g](i) = \sum_{c=1}^C \sum_{h=1}^H \sum_{w=1}^{W_g} F_a((i + w) \% W_a, h, c) \cdot F_g(w, h, c)
\end{equation}

Where \( F_a \in \mathbb{R}^{W_a \times H \times C} \) and \( F_g \in \mathbb{R}^{W_g \times H \times C} \) are the feature maps extracted from aerial and ground images, respectively. \( W_a \) and \( W_g \) denote their widths, while \( H \) denotes the height and \( C \) the number of channels. The modulo operation \( \% \) ensures continuous alignment in 360° images by wrapping column indices that exceed the width of the aerial feature map.

\subsection{Loss Function}
\label{loss_section}
To optimize the parameters of the model during the training phase, we employed the weighted soft-margin triplet loss, expressed as:

\begin{equation}
\mathcal{L} = \log\left(1 + e^{\alpha\left(\|F_g - F_{a'}\|_F - \|F_g - F_{a^{*'}}\|_F\right)}\right)
\end{equation}

Where \( F_g \) represents the unified features of the query ground image, while \( F_{a'} \) and \( F_{a^{*'}} \) correspond to the cropped aerial unified features derived from the matching and non-matching aerial images, respectively. The symbol \( \|\cdot \|_{F} \) denotes the Frobenius norm. The parameter \( \alpha \) regulates the convergence speed of the training process. Furthermore, our approach computes the loss symmetrically: we first use satellite images as queries, treating the ground-level images as positive samples for the true match and negative samples for all other images in the batch. We then reverse this setup. The overall loss is averaged from these two components. This symmetrical computation enables the model to effectively structure the final feature space by maximizing the separation between the query and incorrect matches, while also ensuring that the distance between different queries is maximized, thus encouraging uniqueness among queries in the feature space \cite{Sample4Geo, symmetry}.

\section{Experiment Setup} 
\label{experiment_setup}

\subsection{Implementation Details}
A VGG16 network is employed as a feature extractor for each branch of the architecture. The original VGG16 architecture\cite{simonyan2015deepconvolutionalnetworkslargescale} was modified by removing the last 3 fully connected layers and slightly adapting the remaining 13 convolutional layers. The first 10 layers were kept unchanged while the final 3 convolutional layers were adjusted to alter their stride and padding configurations. These modified layers employ a kernel of size $3 \times 3$ with padding set to \textit{valid} while their stride is set to $(2, 1)$ for the first two of these layers, and  $(1, 1)$ for the last one. The first 10 convolutional layers are initialized with weights pre-trained on ImageNet1K \cite{Imagenet}, while the last 3 layers are randomly initialized. Instead of training the entire network, we froze the first 7 layers to preserve the features learned from ImageNet, fine-tuning only the remaining 6 layers. The Adam optimizer\cite{adam} with a learning rate \(10^{-5}\) is employed for training. The model is trained for 30 epochs using a batch size of dimension 32 through the gradient accumulation technique.

Each input image is resized to a shape of $512 \times 128$. Ground images are also randomly reoriented and, for FoV values below 360°, cropped to the right width. In particular, the FoV value could be \{360°, 180°, 90°, 70°\}, corresponding to an image width of \{512, 256, 128, 99\}. After the features extraction step, the resulting feature map shape is $64 \times 4 \times 16$ for the satellite RGB branch and $64 \times 4 \times 8$ for the satellite semantic segmentation branch. For ground images, the dimensions are $W_g \times 4 \times 16$ for the RGB branch and $W_g \times 4 \times 8$ for the semantic segmentation branch, where depending on the previously mentioned FoV value could be \{64, 32, 16, 12\}.

Assigning more channels to the RGB branches compared to the semantic segmentation branches proved to be effective. While semantic segmentation masks provide valuable contextual information during the matching process, the RGB images remain the primary source of detail, as they more effectively capture essential features like colors and shapes. Following this reasoning, we assigned 16 channels to the RGB branches and 8 channels to the semantic segmentation branches.

%Each input image is reshaped to a size of $512 \times 128$, additionally, the ground images are randomly reoriented and, in the case of a FoV value lower than 360°, cropped to the right width. This results in a final representation of size $64 \times 4 \times C$ for each branch, with C being the number of channels. Concerning this point, giving more importance to the features obtained from the RGB images by assigning more channels to their representation proved effective. A configuration of 16 channels for the RGB and 8 for the semantic segmentation mask provided the best results while semantic segmentation masks provide additional context in the matching phase, the RGB image is still the most informative since most concepts such as colors and shapes are better represented.
As demonstrated in Table \ref{ablation_table2}, and discussed in Section \ref{ablation2}, the proposed combination strategy is more effective than the one used by\cite{pro2024},

%The final number of channels using the proposed combination strategy is 16 and proved to be more effective than one obtained through concatenation, given the same number of channels, as visible from  Table \ref{ablation_table2}.

The evaluation metric used to evaluate this work is the top-K recall with $K=1,5,10$ and $1\%$, corresponding to $K=23$ on the test set, which are reported respectively as r@1, r@5, r@10, and r@1\%. This metric measures how often a ground-view image has its corresponding aerial image within the closest K matches among the whole test set, in terms of distance between their representations. The most significant metric for this task is r@1, as it quantifies the proportion of exact matches achieved. To support reproducibility the code is made publicly available\footnote{https://github.com/MatteoPannacci/SemanticAlignNet-QUAD}.

\subsection{Dataset}

SAN-QUAD was trained and evaluated on an enriched version of the dataset in\cite{pro2024}, a subset of the CVUSA dataset, consisting of 6,647 samples for the training and 2,215 for the evaluation. Each sample comprises a tuple of 4 elements: the ground and satellite RGB images and their respective semantic segmentation masks. The image size is $1232 \times 224$ pixels for the ground images and $370 \times 370$ pixels for the satellite images.

\section{Results} \label{results}

To highlight the effectiveness of SAN-QUAD we compared it against previous state-of-the-art models, on the same subset. Additionally, we conducted several ablation studies to analyze and justify the impact of different components and design choices, providing a comprehensive understanding of the model’s strengths and contributions.

\subsection{Comparison with state-of-the-art} \label{comparison}
To demonstrate the effectiveness of SAN-QUAD architecture in capturing the correlations between ground and satellite images, we benchmarked it against three competitive methods trained and tested over the same dataset. Table \ref{results_comparison} reports the performance of all four methods on the reference dataset, where each model is trained and tested on the same FoV. To perform a direct comparison with previous works, the ground images' FoV values taken into account are 360°, 180°, 90°, and 70°. Notably, our approach shows an improvement in r@1 of 9.8\% on the 360° test when compared to\cite{shi2020ilookingatjoint} and of 5.3\% when compared to SAN\cite{pro2024}, reaching a value of $82.35\%$.  %\textcolor{red}{Although our method does not show a substantial advantage over SAN when trained on a 70° FoV, the improvements scale consistently with the FoV used during the training.}

\begin{table*}[ht]
    \centering
    \begin{adjustbox}{width=\linewidth}
        \begin{tabular}{|c|cccc|cccc|cccc|cccc|}
            \hline
            \multirow{2}{*}{Method} & \multicolumn{4}{c|}{Trained / Tested FoV 360°} & \multicolumn{4}{c|}{Trained / Tested FoV 180°} & \multicolumn{4}{c|}{Trained / Tested FoV 90°} & \multicolumn{4}{c|}{Trained / Tested FoV 70°} \\ \cline{2-17}
             & r@1 & r@5 & r@10 & r@1\% & r@1 & r@5 & r@10 & r@1\% & r@1 & r@5 & r@10 & r@1\% & r@1 & r@5 & r@10 & r@1\% \\ \hline
            Shi et al. \cite{shi2020ilookingatjoint} & 72.87\% & 89.62\% & 92.78\% & 96.52\% & 45.28\% & 72.87\% & 81.76\% & 89.57\% & 6.14\% & 15.80\% & 22.93\% & 35.17\% & 2.21\% & 7.31\% & 12.60\% & 20.50\% \\
            SCN \cite{pro2024} & 54.27\% & 75.67\% & 82.98\% & 89.94\% & 22.21\% & 45.51\% & 57.25\% & 69.84\% & 1.76\% & 6.41\% & 10.02\% & 16.93\% & 1.22\% & 4.11\% & 6.77\% & 13.23\% \\ 
            SAN \cite{pro2024} & 77.07\% & 92.14\% & 95.62\% & 97.97\% & 48.49\% & 75.53\% & 84.06\% & 91.24\% & 6.23\% & 16.43\% & 24.20\% & 37.07\% & 3.02\% & \textbf{9.75\%} & \textbf{15.44\%} & 25.82\% \\ 
            SAN-QUAD (our) & \textbf{82.35\%} & \textbf{94.81\%} & \textbf{96.52\%} & \textbf{98.51\%}
            & \textbf{54.36\%} & \textbf{79.00\%} & \textbf{87.18\%} & \textbf{94.18\%}
            & \textbf{6.55\%} & \textbf{19.86\%} & \textbf{27.95\%} & \textbf{41.94\%}
            & \textbf{3.16\%} & 8.89\% & 14.67\% & \textbf{26.09\%} \\
             % & \textbf{57.29\%} & \textbf{81.67\%} & \textbf{88.67\%} & \textbf{94.49\%}
             % & \textbf{25.73\%} & \textbf{45.78\%} & \textbf{56.25\%} & \textbf{67.22\%}
             % & \textbf{17.56\%} & \textbf{35.35\%} & \textbf{44.02\%} & \textbf{54.63\%} \\ 
            \hline
        \end{tabular}
    \end{adjustbox}
    \vspace{3pt}
    \caption{Comparison of the proposed method (SAN-QUAD) with state-of-the-art across different FoVs. Each model is tested on the same FoV that it is trained on.}
    \label{results_comparison}
\end{table*}

\begin{table*}[ht]
    \centering
    \begin{adjustbox}{width=\linewidth} % Adjust width to fit within page margins
        \begin{tabular}{|c|c|cccc|cccc|cccc|cccc|}
            \hline
            \multirow{2}{*}{Trained FoV} & \multirow{2}{*}{Method} & \multicolumn{4}{c|}{Tested FoV 360°} & \multicolumn{4}{c|}{Tested FoV 180°} & \multicolumn{4}{c|}{Tested FoV 90°} & \multicolumn{4}{c|}{Tested FoV 70°} \\ \cline{3-18}
             &  & r@1 & r@5 & r@10 & r@1\% & r@1 & r@5 & r@10 & r@1\% & r@1 & r@5 & r@10 & r@1\% & r@1 & r@5 & r@10 & r@1\% \\ \hline
            \multirow{3}{*}{360°}
             & Shi et al. \cite{shi2020ilookingatjoint}
             & 72.87\% & 89.62\% & 92.78\% & 96.52\% 
             & 44.47\% & 70.74\% & 79.01\% & 87.63\% 
             & 18.10\% & 36.61\% & 45.60\% & 56.48\% 
             & 11.87\% & 25.06\% & 33.00\% & 43.61\% \\
             & SAN \cite{pro2024}
             & 77.07\% & 92.14\% & 95.62\% & 97.97\% 
             & 47.63\% & 75.30\% & 83.43\% & 90.88\% 
             & 18.65\% & 38.92\% & 48.26\% & 60.00\% 
             & 12.28\% & 28.04\% & 36.84\% & 47.40\% \\
             & SAN-QUAD (our) 
             & \textbf{82.35\%} & \textbf{94.81\%} & \textbf{96.52\%} & \textbf{98.51\%}
             & \textbf{57.29\%} & \textbf{81.67\%} & \textbf{88.67\%} & \textbf{94.49\%}
             & \textbf{25.73\%} & \textbf{45.78\%} & \textbf{56.25\%} & \textbf{67.22\%}
             & \textbf{17.56\%} & \textbf{35.35\%} & \textbf{44.02\%} & \textbf{54.63\%}\\ \hline
            \multirow{3}{*}{180°}
             & Shi et al. \cite{shi2020ilookingatjoint}
             & 66.64\% & 85.06\% & 91.11\% & 95.53\% 
             & 45.28\% & 72.87\% & 81.76\% & 89.57\% 
             & 20.77\% & 41.53\% & 52.42\% & 64.11\% 
             & 13.23\% & 30.20\% & 39.77\% & 52.55\% \\
             & SAN \cite{pro2024}
             & 67.67\% & 86.95\% & 92.46\% & 96.12\% 
             & 48.49\% & 75.53\% & 84.06\% & 91.24\% 
             & 21.08\% & 43.21\% & 53.50\% & 65.91\% 
             & 14.67\% & 30.84\% & 39.82\% & 52.55\% \\ 
             & SAN-QUAD (our) 
             & \textbf{71.69\%} & \textbf{90.11\%} & \textbf{94.40\%} & \textbf{97.83\%} 
             & \textbf{54.36\%} & \textbf{79.00\%} & \textbf{87.18\%} & \textbf{94.18\%} 
             & \textbf{24.10\%} & \textbf{46.82\%} & \textbf{56.88\%} & \textbf{70.52\%}
             & \textbf{15.94\%} & \textbf{33.32\%} & \textbf{44.06\%} & \textbf{56.30\%}\\ \hline
            \multirow{3}{*}{90°}
             & Shi et al. \cite{shi2020ilookingatjoint}
             & 23.48\% & 43.57\% & 53.41\% & 66.59\% 
             & 12.42\% & 29.21\% & 39.19\% & 54.99\% 
             & 6.14\% & 15.80\% & 22.93\% & 35.17\% 
             & 3.61\% & 11.47\% & 18.06\% & 28.04\% \\
             & SAN \cite{pro2024}
             & 23.07\% & 45.96\% & 56.75\% & 70.16\% 
             & 13.72\% & 30.97\% & 42.80\% & 56.88\% 
             & 6.23\% & 16.43\% & 24.20\% & 37.07\% 
             & 3.79\% & 12.10\% & 18.87\% & 29.75\% \\ 
             & SAN-QUAD (our) 
             & \textbf{25.10\%} & \textbf{48.03\%} & \textbf{59.86\%} & \textbf{73.05\%}
             & \textbf{13.95\%} & \textbf{34.04\%} & \textbf{45.01\%} & \textbf{61.85\%} 
             & \textbf{6.55\%} & \textbf{19.86\%} & \textbf{27.95\%} & \textbf{41.94\%}
             & \textbf{4.10\%} & \textbf{13.95\%} & \textbf{22.08\%} & \textbf{33.63\%} \\ \hline
            
            \multirow{3}{*}{70°}
             & Shi et al. \cite{shi2020ilookingatjoint}
             & 11.24\% & 26.37\% & 35.85\% & 49.93\% 
             & 6.46\% & 15.58\% & 23.34\% & 36.07\% 
             & 2.84\% & 9.57\% & 14.09\% & 24.74\% 
             & 2.21\% & 7.31\% & 12.60\% & 20.50\% \\
             & SAN \cite{pro2024}
             & \textbf{14.67\%} & \textbf{33.18\%} & \textbf{42.93\%} & 56.52\%
             & \textbf{7.27\%} & \textbf{21.40\%} & \textbf{30.43\%} & 44.20\% 
             & 3.43\% & \textbf{12.19\%} & \textbf{19.82\%} & \textbf{31.24\%}
             & 3.02\% &\textbf{9.75\%} & \textbf{15.44\%} & 25.82\% \\ 
             & SAN-QUAD (our) 
             & 13.32\% & 30.47\% & 41.94\% & \textbf{56.75\%}
             & 6.68\% & 19.86\% & 29.57\% & \textbf{44.65\%}
             & \textbf{3.57\%} &10.93\% & 17.38\% & 29.71\% 
             & \textbf{3.16\%} & 8.89\% & 14.67\% & \textbf{26.09\%} \\ \hline
        \end{tabular}
    \end{adjustbox}
    \vspace{3pt}
    \caption{Comparison of the proposed method (SAN-QUAD) with state-of-the-art across different FoVs. Each method is trained on a specific FoV and tested on all of them.}
    \label{full_comparison_table}
\end{table*}

%%%%%CHECH THIS! I THINK COULD BE BETTER! NEW!

To prove the generalization capabilities of the model, the architectures are tested on FoVs different from those used during training, results are reported in Table \ref{full_comparison_table}. Previous works have shown that models trained on a 360° field of view perform worse on narrower test FoVs with respect to their counterparts trained on 180°. Additionally, models trained on 90° and 70° FoVs exhibit a significant drop in performance across all test FoVs. SAN-QUAD presents a slightly different behavior: the model trained on the 360° FoV outperforms the one trained on 180° in r@1, the most relevant metric for the task. We can observe a significant reduction in the performance gap on narrower FoVs differently from previous works.

% This behavior is inconsistent across all metrics, but we can observe a significant reduction in the performance gap on narrower FoVs with respect to previous works.

%%%%%%%%%%%%% COMPARE TO THIS! OLD!

% In previous works, models trained on a 360° FoV never outperformed those trained on narrower FoVs when tested on limited fields of view, \textcolor{red}{in particular the models trained on 180° FoV perform better on them while those trained on even lower FoVs have a drastic decrease in performances}. Instead, using our proposed method, the model trained on the 360° FoV not only outperforms all its other versions across all considered FoVs but also all tested benchmarks on their best results. Our approach achieves an improvement of 10\% on r@1 for the 360° test when compared to Shi et al. and 5\% compared to SAN. While the proposed method does not show a significant advantage over SAN when trained on the 70° FoV, it delivers notable improvements when tested on it using models trained on larger FoVs.

Training on larger FoVs allows the ground semantic segmentation branch of SAN-QUAD to learn better the patterns and structures within the segmentation masks. These learned features remain relevant even when tested on smaller portions of the images, enabling the model to generalize effectively. Training instead on narrower FoVs provides limited context, restricting the branch’s ability to extract meaningful features, which can reduce its contribution to the overall performance.

% present all the models used in the comparison

% comparison of full model and "experiments" with different FoVs

% \begin{table*}[ht]
%     \centering
%     \begin{tabular}{ |c|cccc| } 
%         \hline
%         \textbf{Models} & \textbf{r@1} & \textbf{r@5} & \textbf{r@10} & \textbf{r@1\%} \\
%         \hline
%         Shi et al. \cite{shi2020ilookingatjoint} & 72.87\% & 89.62\% & 92.78\% & 96.52\% \\
%         SCN \cite{pro2024} & 54.27\% & 75.67\% & 82.98\% & 89.94\% \\
%         SAN \cite{pro2024} & 77.07\% & 92.14\% & 95.62\% & 97.97\% \\
%         SAN-QUAD (our) & \textbf{82.35\%} & \textbf{94.81\%} & \textbf{96.52\%} & \textbf{98.51\%} \\
%         \hline
%     \end{tabular}
%     \vspace{3pt}
%     \caption{Comparison of the proposed method (SAN-QUAD) with state-of-the-art. The models are trained and tested on 360° ground images.}
%     \label{merged_table}
% \end{table*}

\subsection{Ablation Study} \label{ablation}

To further justify the adopted implementation choices and the discussed intuitions about the model behavior, three additional experiments were conducted as an ablation study.

\subsubsection{Semantic stream evaluation} \label{ablation1}

The first experiment aimed to prove the effect of including additional semantic features. We considered the model proposed by Shi et al.\cite{shi2020ilookingatjoint}, which uses only the RGB ground and satellite images, as the baseline and presents models that incrementally include additional features to assess their impact. 

Results are reported in Table \ref{ablation_table1}. The TRIPLE-SAT model, adapted from the SAN method in \cite{pro2024}, incorporates our proposed features combination technique to ensure a fair basis for comparison. This modification enables us to attribute performance differences solely to the introduced features. The TRIPLE-GRD model is an analogous three-branch architecture, but it includes the ground semantic segmentation branch instead of the satellite one. Finally, the SAN-QUAD model is the four-branch architecture proposed in this work.

Comparing the two models that use either ground or satellite semantic segmentation, is visible that that the former consistently outperforms the latter, suggesting that the ground semantic segmentation masks may be more informative than the satellite ones. Integrating both features, through the SAN-QUAD architecture, results in the best performances, outperforming the usage of both ground and satellite segmentation individually. This result highlights the importance of combining features from both viewpoints to improve performance.

\begin{table}[ht]
    \centering
    \begin{adjustbox}{width=\linewidth}
    \begin{tabular}{ |c|cccc| }
        \hline
        \textbf{Models} & \textbf{r@1} & \textbf{r@5} & \textbf{r@10} & \textbf{r@1\%} \\
        \hline 
        Shi et al. \cite{shi2020ilookingatjoint} & 72.87\% & 89.62\% & 92.78\% & 96.52\% \\
        TRIPLE-SAT & 78.60\% & 92.91\% & 96.07\% & 98.42\% \\
        TRIPLE-GRD & 80.50\% & 93.63\% & 96.03\% & 98.15\% \\
        SAN-QUAD & 82.35\% & 94.81\% & 96.52\% & 98.51\% \\
        \hline
    \end{tabular}
    \end{adjustbox}
    \caption{Ablation Study 1: Effect of the inclusion of features.}
    \label{ablation_table1}
\end{table}

\subsubsection{Combination strategy} \label{ablation2}

The second experiment compared the two feature combination strategies discussed in this work. The results, presented in Table \ref{ablation_table2}, demonstrate that using the partial sum operator for combining the features of the branches leads to improved performance compared to the concatenation technique. This is consistent for both the proposed SAN-QUAD architecture and the original SAN architecture.

The table also shows that SAN-QUAD with the concatenation operator performs only marginally better than SAN. As discussed in Section \ref{comb_strategies}, this may be due to the matching process, where each feature is compared only with those in the same channel from the opposite viewpoint, while introducing shared features across branches mitigates this limitation.

\begin{table}[ht]
    \centering
    \begin{adjustbox}{width=\linewidth}
    \begin{tabular}{ |c|cccc| }
        \hline
        \textbf{Models} & \textbf{r@1} & \textbf{r@5} & \textbf{r@10} & \textbf{r@1\%} \\
        \hline 
        SAN (concat.) \cite{pro2024} & 77.07\% & 92.14\% & 95.62\% & 97.97\% \\
        SAN (sum) & 78.60\% & 92.91\% & 96.07\% & 98.42\% \\   
        \hline
        SAN-QUAD (concat.) & 78.24\% & 91.87\% & 95.21\% & 97.79\% \\        
        SAN-QUAD (sum) & 82.35\% & 94.81\% & 96.52\% & 98.51\% \\
        \hline
    \end{tabular}
    \end{adjustbox}
    \caption{Ablation Study 2: Effect of the combination strategy.}
    \label{ablation_table2}
\end{table}

\subsubsection{Depth estimation} \label{ablation3}

In the third experiment, we evaluated the inclusion of an additional feature, the depth map of the ground-view images, adding a fifth stream depicted to extract features from the new information. So, we compare our proposed model SAN-QUAD with a five-branches variant, named the QUINTUPLE model. The adopted combination strategy is the one presented in this work, extended to deal with the additional branch. To generate the depth maps, the Depth-Anything model \cite{yang2024depthanythingunleashingpower} was used.

The results show a decrease in performance compared to the proposed architecture. Our intuition is that the depth estimation is incompatible with a late-fusion strategy. %, as it's difficult to extract good features from that alone. Instead adopting an early-fusion strategy could get better results as the network could learn how to combine the features more effectively. 
Quantitative results are reported in Table \ref{ablation_table3}.

\begin{table}[ht]
    \centering
    \begin{adjustbox}{width=\linewidth}
    \begin{tabular}{ |c|cccc| }
        \hline
        \textbf{Models} & \textbf{r@1} & \textbf{r@5} & \textbf{r@10} & \textbf{r@1\%} \\
        \hline 
        SAN-QUAD & 82.35\% & 94.81\% & 96.52\% & 98.51\% \\
        QUINTUPLE & 75.62\% & 91.69\% & 95.62\% & 97.79\% \\
        \hline
    \end{tabular}
    \end{adjustbox}
    \caption{Ablation Study 3: Effect of the depth estimation stream.}
    \label{ablation_table3}
\end{table}

\section{Conclusion}

This work introduces SAN-QUAD, an architecture to address the problem of geo-locating ground-view images and matching them with aerial images, without relying on GPS data. The used approach incorporates information extracted from semantic segmentation masks alongside RGB images from both ground and satellite perspectives which are then merged, resulting in significant performance improvement over existing models compared on a subset of the CVUSA dataset.

This system can be applied as a support tool in forensic analysis, particularly in cases where it is necessary to reconstruct the location of a non-geo-tagged image or where geolocating an image could aid in verifying the authenticity of reported news.

In future works, we will integrate additional features for ground and satellite views, considering combining them before passing through the feature extractors. Moreover, we plan on generalizing the sum operation by considering a learnable combination of feature channels to build the final representation of each viewpoint. We also plan to extend the work to other datasets to train and evaluate the proposed method.

\section{Acknowledgment}
This study has been partially supported by SERICS (PE00000014) under the MUR National Recovery and Resilience Plan funded by the European Union - NextGenerationEU.
%%%%%%%%% REFERENCES
{\small
\bibliographystyle{ieee_fullname}
\bibliography{egbib}
}

\end{document}